\documentclass[journal]{IEEEtranTIE}
\usepackage{graphicx}
\usepackage{cite}
\usepackage{picinpar}
\usepackage{amsmath}
\usepackage{url}
\usepackage{flushend}
\usepackage[latin1]{inputenc}
\usepackage{colortbl}
\usepackage{soul}
\usepackage{multirow}
\usepackage{pifont}
\usepackage{color}
\usepackage{alltt}
\usepackage[hidelinks]{hyperref}
\usepackage{enumerate}
\usepackage{siunitx}
\usepackage{breakurl}
\usepackage{epstopdf}
\usepackage{pbox}

\begin{document}
\title{	Robust Control of An Aerial Manipulator Based on A Variable Inertia Parameters Model}

\author{
	\vskip 1em
	{
	Guangyu Zhang,
	Yuqing He, \emph{Membership},
  Bo Dai, 
  Feng Gu,
  Jianda Han, \emph{Membership},
  Guangjun Liu, \emph{Senior Membership}
	}

	\thanks{
		
		{
		Manuscript received June 3, 2019; revised October 20, 2019; accepted November 8, 2019.
    This work was supported in part by the National Nature Sciences Foundation of China, grant number 61433016 and U1608253.
		
    G. Zhang, Y. He, B. Dai, F. Gu and J. Han are with the State Key Laboratory of Robotics, Shenyang Institute of Automation, Chinese Academy of
    Sciences, Shenyang 110016, China, and also with Institute for Robotics and Intelligent Manufacturing, Chinese Academy of
    Sciences, Shenyang 110016, China(e-mail: zhangguangyu15@mails.ucas.edu.cn; heyuqing@sia.cn; daibo16@mails.ucas.edu.cn; fenggu@sia.cn; jdhan@sia.cn).
    
    G. Liu is with the Department of Aerospace Engineering, Ryerson University, Toronto, ON M5B 2K3 Canada(e-mail: gjliu@ryerson.ca).
    }
	}
}

\maketitle
	
\begin{abstract}
Aerial manipulator, which is composed of an UAV (Unmanned Aerial Vehicle) and a multi-link manipulator and can perform aerial manipulation, has shown great potential of applications. However, dynamic coupling between the UAV and the manipulator makes it difficult to control the aerial manipulator with high performance. In this paper, system modeling and control problem of the aerial manipulator are studied. Firstly, an UAV dynamic model is proposed with consideration of the dynamic coupling from an attached manipulator, which is treated as disturbance for the UAV. In the dynamic model, the disturbance is affected by the variable inertia parameters of the aerial manipulator system. Then, based on the proposed dynamic model, a disturbance compensation robust $H_{\infty}$ controller is designed to stabilize flight of the UAV while the manipulator is in operation. Finally, experiments are conducted and the experimental results demonstrate the feasibility and validity of the proposed control scheme.
\end{abstract}


\markboth{IEEE TRANSACTIONS ON INDUSTRIAL ELECTRONICS (Accepted Version)}%
{}

\definecolor{limegreen}{rgb}{0.2, 0.8, 0.2}
\definecolor{forestgreen}{rgb}{0.13, 0.55, 0.13}
\definecolor{greenhtml}{rgb}{0.0, 0.5, 0.0}

\section*{Nomenclature}
\begin{table}[h]  \small
\begin{tabular}{ll}
$\Sigma_B$ & UAV body fixed coordinate frame. \\
$O$ & The coordinate origin point of $\Sigma_B$. \\
$\Sigma_I$ & Inertial coordinate frame. \\
$\Sigma_j$ & Manipulator's  joint  frames. $j=1,2,3,\dots, m$.   \\
$r_{op}$ &  Vector from $O$ to point $p$ with respect to $\Sigma_I$.\\
${}^Br_{op}$ & Vector from $O$ to point $p$ with respect to $\Sigma_B$. \\
${}^B\omega_b$ & Angular velocity of the UAV with respect to $\Sigma_B$. \\
$m_{man}$ & Mass of manipulator. \\
$m_b$ & Mass of UAV. \\
$m_s$ & Mass of total aerial manipulator system.\\
$C$, $C_m$ & \multicolumn{1}{p{6.5cm}} {CoM (Center of mass) of total aerial manipulator system and manipulator, respectively.} \\
$r_{oc}$ & Vector from $O$ to $C$ in with respect to $\Sigma_I$. \\
${}^Br_{oc}$ & Vector from $O$ to $C$ with respect to $\Sigma_B$. \\
$P_I$, $L_I$ & Linear and angular momentum of the system. \\
${}^IR_B$ & Rotation matrix from $\Sigma_B$ to $\Sigma_I$. \\
\end{tabular}
\end{table}
\begin{table}[h]  \small
\begin{tabular}{ll}
${}^BI_b^o$ & \multicolumn{1}{p{6.5cm}} {Inertia matrix of the UAV referenced to the point $O$ with respect to $\Sigma_B$.} \\
${}^BI_{man}^o$ & \multicolumn{1}{p{6.5cm}} {Inertia matrix of the manipulator referenced to the point $O$ with respect to $\Sigma_B$.} \\
${}^BL_{man}$ & \multicolumn{1}{p{6.5cm}} {Angular momentum of the manipulator with respect to $\Sigma_B$.} \\
${}^Br_{ocm}$ & Vector from point $O$ to $C_m$ with respect $\Sigma_B$.\\
${}^IF_{ext}$ & External force acting on system with respect to $\Sigma_I$. \\
${}^IM_{ext}^o$ & External moment acting on point $O$ with respect to $\Sigma_I$. \\
${}^jp_{cj}$ & Position of the CoM of link $j$ with respect to $\Sigma_j$.\\
${}^Bv_{cj}$ & Velocity of the CoM of link $j$ with respect to $\Sigma_B$. \\
${}^B\omega_{j}$ & Angular velocity of link $j$ with respect to $\Sigma_B$. \\
${}^BR_j$ & Rotation matrix from $\Sigma_j$ to $\Sigma_B$. \\
${}^BJ_{cj}$ & Jacobian matrix of the CoM of link $j$ with respect $\Sigma_B$. \\
$m_j$ & Mass of manipulator link $j$. \\
$I_j^{cj}$ & Inertia of the link $j$ referenced to its CoM. \\
$p_b$, $v_b$ & Position and velocity of UAV with respect to $\Sigma_I$.\\
$\Phi_b$ & Euler angles of UAV including roll, pitch and yaw. \\
$I_b$ & UAV inertial matrix referended to $O$. \\
$F_{dis}$ & Force disturbance of manipulator with respect to $\Sigma_I$. \\
${}^B\tau_{dis}$ & Torque disturbance of manipulator with respect to $\Sigma_B$. \\
$\hat{F}_{dis}$ & Estimated force disturbance. \\
${}^B\hat{\tau}_{dis}$ & Estimated torque disturbance.\\
$\Delta F$, $\Delta \tau$ & Force and torque disturbance residual error.\\
\end{tabular}
\end{table}

\section{Introduction}

\IEEEPARstart{A}{ERIAL} manipulation has become a new research hotspot in the field of aerial robotics, in recent years \cite{khamseh2018aerial,lee2017estimation}. An aerial manipulator is typically composed of a rotor craft UAV (e.g., helicopter or multi-rotor) and a multi-link robotic arm. Such an aerial robot can extend applications of the UAV from passive scenarios, e.g. exploration and surveillance, to active scenarios, e.g. grasping and manipulating. For example, in \cite{thomas2014toward} an aerial manipulator was used to grasp a static tubular object, mimicking an agile that captures a prey.
When a manipulator is mounted on a rotor craft UAV, the dynamics of the two systems are strongly coupled, which makes it challenging to precisely control their movement. From control perspective, the reported works in the aerial manipulator model and control can be roughly divided into two categories. For the first category, the whole aerial manipulator system is considered as one controlled object, and one controller is designed to stabilize the states of UAV and manipulator simultaneously. For the second category, a separate control strategy is used,  which means that two separated controllers are designed for the UAV and manipulator, respectively. The dynamic coupling between them is considered as external disturbance for both UAV sub-system and manipulator sub-system.

Research works reported in \cite{lippiello2012cartesian,lippiello2012exploiting,kim2016vision,yang2014dynamics,kobilarov2014nonlinear,caccavale_adaptive_2014,arleo_control_2013} are all belong to the first category. In \cite{lippiello2012cartesian} and \cite{lippiello2012exploiting}, the dynamic model of an aerial manipulator with 6+n DoF (Degree of Freedom) was derived through Euler-Lagrange equation. Based on the dynamic model, a Cartesian impedance controller was designed to realize hovering of the UAV when the manipulator was contacting with the environment. In grasping tasks, to make the aerial manipulator robust against disturbances due to contact forces, reference \cite{kim2016vision} proposed a passivity based adaptive controller using the same dynamic model. 
One obvious property of the aerial manipulator different from the traditional manipulator is system's under-actuation, because of the under-actuation property of the UAV. Some research has taken the under-actuation of the aerial manipulator into full consideration in the controller design. Such as, in \cite{yang2014dynamics}, the dynamic model of a quadrotor manipulator was decoupled into translational dynamics of system's CoM, an under-actuated subsystem, and rotational dynamics of the quadrotor and manipulator, a fully-actuated subsystem. Based on this structure, a back-stepping like controller was designed to assign different roles for its CoM and end-effector control. A similar decoupling method and control strategy were also presented in \cite{kobilarov2014nonlinear}.
In \cite{caccavale_adaptive_2014,arleo_control_2013}, the hierarchical inner-outer loop
control scheme, which can handle the UAV's under-actuation\cite{kendoul2008asymptotic},
is used in an adaptive controller of aerial manipulator.
The rigid dynamics of the aerial manipulator is a large-dimensional 
nonlinear and under-actuated system. Hence a controller based the overall model of the 
aerial manipulator dynamics is so complicated that it is difficult to implement onboard. 
In order to make the controller implement onboard more easily, extensive research has been 
conducted based on the second control strategy.

For the second category, most works are focused on the UAV control, as steady flight of the UAV is essential for manipulation tasks. When the UAV controller is designed, the coupling effect of the manipulator is treated as disturbance. Hence the study of the coupling effect of the manipulator on the UAV dynamics is of crucial importance. There have been two different ways to deal with the coupling effect. 

\begin{enumerate}[1)]
    \item   Coupling effect is represented by the interaction 
    force/torque between the UAV and the manipulator. To obtain 
    the interaction force/torque, in \cite{kondak2013closed}, 
    a force/torque sensor was installed between a helicopter and 
    a 7-DoF manipulator. In \cite{kim2018robust}, a disturbance 
    observer (DoB) is used in UAV controller to estimate the 
    interaction force/torque which is seen as disturbance of the 
    UAV. The recursive Newton-Euler method 
    was used in the aerial manipulator dynamic 
    modeling in \cite{khalifa2017new} and \cite{fanni2017new}. 
    The experimental results have shown that the 
    control performance of UAV can be improved. However 
    if they can use more state information of the moving 
    manipulator to estimate the interaction force more accurate, 
    the results will be better.
    
    \item   Coupling effect is described by inertia parameters (CoM and inertia matrix). The aerial manipulator can be taken as a special aerial platform whose mass distribution could be changed due to the movement of the manipulator. 
    The motion of the manipulator can be represented by 
    the increment of the CoM and the inertia matrix, which are varying 
    when the manipulator is moving.
    So they are called as variable inertia parameters. 
    The variable inertia parameters in the UAV's dynamic model can reflect the disturbance of the moving manipulator, as in \cite{jimenez2016modelling,orsag2017dexterous,zhang2018grasp}. In order to reject the disturbance, the variable parameters are compensated through the inverse dynamics in the controller design. The variable inertia parameters in \cite{jimenez2016modelling,orsag2017dexterous,zhang2018grasp} can only compensate the gravity of the manipulator, so it is not suitable when the manipulator is moving quickly.
\end{enumerate}

Aiming at hovering manipulation tasks, this paper is mainly focused on the UAV dynamic modelling and control under the coupling effect of the manipulator. The main contributions of this article are as follows: 

\begin{enumerate}[1)]
\item	A dynamic model of an UAV with a manipulator attached is derived using variable inertia parameters to describe the coupling effect. The coupling effect model includes the first-order and second-order derivatives of the variable inertia parameters, which means that it is suitable when the manipulator is moving quickly, e.g. fast grasp task. To the best of our knowledge, only static effects have been considered in reported research works.

\item	A disturbance compensation robust $H_{\infty}$ controller is designed to stabilize the UAV while the manipulator is moving. The controller is composed of a coupling effect model based disturbance estimator and a robust $H_{\infty}$ compensator, which can compensate the disturbance of manipulator effectivly.
\end{enumerate}

The rest of this paper is organized as follows. In Section II, the aerial manipulator dynamic model is presented, which uses the variable inertia parameters to describe the coupling effect. The control scheme is proposed in Section III. Subsequently, the experiments and results are given in Section IV. Finally, conclusions are included in Section V.

\section{Dynamic model of the aerial manipulator}
As we know, aerial vehicle is typically modeled as a six DoF rigid body, and its dynamics can be derived using the Newton-Euler method. When the aerial manipulator is taken as a special aerial platform whose mass distribution is changing, instead of a rigid body, we can use the Linear and Angular Momentum Theorem of the mass point system to obtain its dynamic model \cite{wittenburg2007dynamics}.

\subsection{Momentum of the aerial manipulator }
\begin{figure}[!t]
    \centering
    \includegraphics[width=7cm]{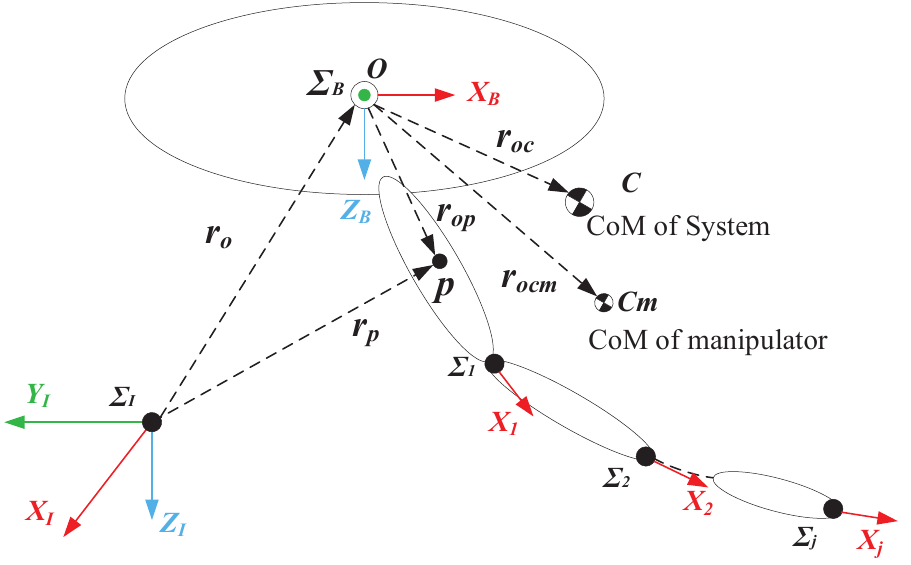}\vspace{-6pt}
	\caption{Rigid body frame of the aerial manipulator.}\label{fig_1}
	\vspace{-6pt}
\end{figure}

As shown in \mbox{Fig. \ref{fig_1}}, we use several ellipsoids to denote the aerial manipulator, where the largest ellipsoid denotes the UAV (multi-rotor or helicopter) and the other ellipsoids denote the links of the manipulator. Let $\Sigma_I$ denote the NED (North-East-Down) inertial coordinate frame. $\Sigma_B$ denotes the UAV body fixed coordinate frame with its origin at point $O$, which is the CoM of the UAV. The $X_B$ and $Y_B$ axes of the $\Sigma_B$ are in the directions of the head and right of the UAV, respectively. The manipulator's joint frames are constructed based on standard DH parameters and denoted by $\Sigma_j$ ( $j=1,2,3,\cdots, n$ ).

Suppose that point $p$ is an arbitrary mass point of the aerial manipulator, as shown in \mbox{Fig. \ref{fig_1}}. The vector $r_p$ and $r_o$ are the absolute position of point $p$ and point $O$ with respect to $\Sigma_I$, respectively. $r_{op}$ is the vector from point $O$ to point $p$. The relationship among $r_{op}$, $r_p$ and $r_o$ is as follow:
\begin{equation} 
{r_p} = {r_o} + {r_{op}} \label{eq1} 
\end{equation}
If we suppose that ${^B} r_{op}$ is the indication of $r_{op}$ with respect to $\Sigma _B$. $^I\textrm{} R_B$ is the rotation matrix from $\Sigma_B$ to $\Sigma_I$, so,
\begin{equation}
    {r_{op}} = {}^I{R_B}(^B{r_{op}}) \label{eq2}
\end{equation}
Then we have, 
 \begin{equation}
    {\dot r_{op}} = {}^IR_B ({}^B{\omega _b} \times ^B{r_{op}} + ^B{\dot r_{op}}) \label{eq3}
\end{equation}
where, $^B \omega_b$ is the angular velocity of the UAV with respect to $\Sigma_B$. 

As shown in \mbox{Fig. \ref{fig_1}}, the CoM of the whole aerial manipulator system is denoted by point $C$. $r_{oc}$ is the vector from point $O$ to $C$, which has a relationship with $r_{op}$ as the following equation:
\begin{equation}
    \int_{m_b + m_{man}} r_{op} { }\text{d}m_p=(m_b + m_{man}) r_{oc} = m_s r_{oc}
    \label{eq4}
\end{equation}
where, $m_b$ and $m_{man}$ are the mass of the UAV and the manipulator, respectively. $m_s$ is the total mass of the aerial manipulator system, that is $m_s = m_b + m_{man}$. $\text{d}m_p$ is the mass of point $p$.

For the aerial manipulator system, its absolute linear and angular momentum, denoted by $P_I$ and $L_I$, are defined by the following equations:
\begin{equation}
     \begin{cases}
       P_I = \int_{m_b + m_{man}} \dot r_p \text{d} m_p\\
       L_I = \int_{m_b + m_{man}} r_p \times \dot r_p \text{d} m_p\\
     \end{cases} \label{eq5}
\end{equation}
Combining with (\ref{eq1})-(\ref{eq5}), we can get the absolute linear and angular momentum of the multi-body system as follows: 
 \begin{equation}
   P_I = m_s(\dot r_o + ^IR_B(^B{\omega _b} \times ^B{r_{oc}} + ^B{\dot r_{oc}}))
   \label{eq6}
 \end{equation} 
\begin{equation}
    \begin{aligned}
      L_I = & r_o \times P_I + m_s r_{oc} \times \dot r_o  \\
            &+ ^IR_B (({^B I^o_b} {^B{\omega}_b} + {^B I^o_{man}} {^B {\omega} _b} )+{}^B{L_{man}})\\
    \end{aligned}
    \label{eq7}
\end{equation}
where, $^B{r_{oc}}$ is the indication of $r_{oc}$ with respect to $\Sigma _B$.  $^B{ \dot  r_{oc}}$ is the velocity of the point $C$ relative to point $O$ with respect to $\Sigma _B$.  $^B I_b^o$ and $^B I_{man}^o$  are the inertia matrix of the UAV and the manipulator referenced to the point $O$ along to body fixed frame axes with respect to $\Sigma _B$, respectively.  $^B{L_{man}}$ is the angular momentum of manipulator with respect to $\Sigma _B$, and it is defined as:
${}^B{L_{man}} = \int_{{m_{man}}} {({}^B{r_{op}} \times {}^B{{\dot r}_{op}})d{m_p}}$

As shown in \mbox{Fig. \ref{fig_1}}, the CoM of the manipulator is denoted by point $Cm$. ${}^B{r_{ocm}}$ is the vector form point $O$ to point $Cm$ with respect to $\Sigma _B$. In order to simplify $^B{L_{man}}$, we use the angular momentum of the manipulator's CoM relative to the point $O$ to approximate it, as follow: 
\begin{equation}
    {}^B{L_{man}} \approx m_{man} {}^B{r_{ocm}} \times {}^B{\dot r_{ocm}} = \frac{m_s^2} {m_{man}} {}^B{r_{oc}} \times {}^B{\dot r_{oc}} \label{eq9}
\end{equation}
where,  ${}^B{\dot r_{ocm}}$ is the velocity of the point $Cm$ relative to point $O$ with respect to $\Sigma _B$. The experiment results in the Section IV show that the main part of the coupling effect is contained after simplifying $^B{L_{man}}$. 

\subsection{Dynamics of the aerial manipulator}
In order to get the dynamics of the aerial manipulator, we can use the Linear and Angular Momentum Theorem\cite{wittenburg2007dynamics}.
It is as follows: 
\begin{equation}
    \begin{cases}
     \dot{P_I} = {}^I{F_{ext}}\\
     \dot{L_I} = {r_o} \times {}^I{F_{ext}} + {}^I M_{ext}^o
    \end{cases} \label{eq10}
\end{equation}
where, $^I F_{ext}$ is the total external fore acting on the system with respect to $\Sigma _I$. ${}^I M_{ext}^o$ is the total moment acting on point $O$ with respect to $\Sigma _I$. For the rotor wing UAV system they can be denoted in detail as follows:
\begin{equation}
    \begin{cases}
    {}^I{F_{ext}} =  - {F_t}{}^IR{}_B{e_3} + {m_s}g{e_3}\\
    {}^I M_{ext}^o = {}^I{R_B}\tau  + {m_s}{}^I{R_B}{}^B{r_{oc}} \times g{e_3}
    \end{cases} \label{eq11}
\end{equation}
where, $F_t$ and $\tau$ are the thrust and moment generated by rotors of the UAV action on the point $O$. $g$ is gravity acceleration. $e_1$, $e_2$ and $e_3$ are unit vectors, which means that the unit matrix ${I_{3 \times 3}} = [{e_1}{\text{ }}{e_2}{\text{ }}{e_3}]$ .

Through derivating (\ref{eq6}) and (\ref{eq7}), $\dot{P_I}$ and $\dot{L_I}$ can be got. Then combining  with (\ref{eq10})-(\ref{eq11}), we can obtain the dynamics of the aerial manipulator as follows:
\begin{equation}
    \begin{split}
        m_s \ddot r_o = & - {F_t}{}^IR{}_B{e_3} + {m_s}g{e_3}   \\
        &  - m_s {}^IR{}_B({}^B{\omega _b} \times ({}^B{\omega _b} \times {}^B{r_{oc}}) + {}^B{{\dot \omega }_b} \times {}^B{r_{oc}})  \\ 
        &  - m_s {}^IR{}_B(2{}^B{\omega _b} \times {}^B{{\dot r}_{oc}} + {}^B{\ddot r}_{oc}) 
    \end{split}\label{eq12}
\end{equation}
\begin{equation}
    \begin{split}
       &({}^BI_b^o + {}^B I_{man}^o){}^B{{\dot \omega}_b} =  \tau-{}^B{\omega _b} \times (({}^BI_b^o + {}^BI_{man}^o){}^B{\omega _b}) \\ 
       &  + {m_s}({}^B{r_{oc}} \times {}^IR_B^{ - 1}(g{e_3} - {{\ddot r}_o})) - {}^B\dot I_{man}^o{}^B{\omega _b}  \\
       &- ({}^B{\omega _b} \times {}^B{L_{man}})- {}^B{{\dot L}_{man}} 
    \end{split}\label{eq13}
\end{equation}

Equation (\ref{eq12}) and (\ref{eq13}) are the dynamics of aerial manipulator under the actuating of rotors' aerodynamic force, so it describes the dynamics of the UAV and contains the effect of the manipulator. In (\ref{eq12}) and (\ref{eq13}), the coupling effect between the UAV and the manipulator is described by the terms including variable inertia parameters, $^B{r_{oc}}$, ${}^B I_{man}^o$ and their derivative. In the rest of this section, we will introduce a algorithm to calculate the variable inertia parameters.

\subsection{Variable inertia parameters}
The position of CoM of link $j$ with respect to $\Sigma _j$ and $\Sigma _B$ are denoted by $^j p_{cj}$ and $^B{p_{cj}}$, respectively. The velocity and angular velocity of CoM of link $j$ with respect to $\Sigma _B$ are denoted by $^B{v_{cj}}$ and $^B \omega_j$, respectively. Let $q$ denote the vector of the joint angle of the manipulator. Based on the kinematics of the manipulator, we can get the following relationship:
\begin{equation}
    \begin{cases}
    {}^B{p_{cj}} = {}^B{R_j}{}^j{p_{cj}}\\
    {\left[ 
      \begin{matrix}
        {{}^B{v_{cj}}}  \\ 
        {{}^B{\omega _j}}  \\ 
      \end{matrix}  \right]} = {}^B{J_{cj}}\dot q
    \end{cases} \label{eq14}
\end{equation}
where, $^B{R_j}$ is the rotation matrix from $\Sigma _j$ to $\Sigma _B$. $^B{J_{cj}}$ is the manipulator Jacobian matrix of the CoM of link $j$.  $\dot q$ is the vector of joint velocity of the manipulator.

So $^B{r_{oc}}$ and its derivative can be calculated by following equations:
\begin{equation}
    \begin{cases}
      {}^B{r_{oc}}=\frac{1}{m_s}\sum\limits_{j = 1}^n {{m_j}{}^B{p_{cj}}} \\
      {}^B{{\dot r}_{oc}}=\frac{1}{m_s}\sum\limits_{j = 1}^n {{m_j}{}^B{v_{cj}}} 
    \end{cases} \label{eq15}
\end{equation}
where, $m_j$ is the mass of the link $j$.

Based on the definition of inertia matrix and parallel axis theorem \cite{wittenburg2007dynamics}, ${}^B I_{man}^o$ and its derivative can be obtained as follows:
\begin{equation}
  \begin{split}
      {}^B{I_{man}^o} = &\sum\limits_{j=1}^n ({}^B{R_j}I_j^{cj}{}^BR_j^{-1} + m_j(\|{}^B p_{cj}{\|^2}{I_{3\times 3}} \\
      & -{}^B p_{cj}{{({}^Bp_{cj})}^T})
 \end{split} \label{eq16}  
\end{equation}
\begin{equation}
    \begin{split}
       & {}^B\dot I_{man}^o = \sum\limits_{j = 1}^n (Skew{(^B}{\omega _j}){}^B{R_j}I_j^{cj}{}^j{R_B} \\
       &- {}^B{R_j}I_j^{cj}{}^j{R_B}Skew{(^B}{\omega _j}))    + \sum\limits_{j = 1}^n {m_j}(2{{({}^B{p_c}_j)}^T}{}^B{v_{cj}}{I_{3 \times 3}} \\
       &- {}^B{v_c}_j{{({}^B{p_{cj}})}^T} - {}^B{p_c}_j{{({}^B{v_{cj}})}^T})  
    \end{split} \label{eq17}
\end{equation}
where, $I^{cj}_j$is the inertia of the link $j$ referenced to its CoM. $Skew(.)$ is skew symmetric matrix function of a vector.
\textbf{Remark I}: The variable inertia parameters are rely only on the joint states and some constants, and all these are obtainable online or offline through directly measurement or estimation. 
\section{Control of the aerial manipulator system }
In the previous section we have presented the rigid body dynamic model of an UAV with a manipulator. In this section we will use the dynamic model to design a controller of an aerial manipulator composed of a hex-rotor and a multi-link manipulator.
\subsection{Dynamics of the Hex-rotor Manipulator}
In order to make the dynamic model more understandable for controller design of the hex-rotor manipulator, we need to rewrite the dynamic model using the state variable of the hex-rotor firstly.

The position and velocity of hex-rotor with respect to $\Sigma _I$ are denoted by 
${p_b} = { \left[ {\begin{matrix} x  & y  & z \end{matrix} } \right]^T}$ 
and 
${v_b} = { \left[ {\begin{matrix} v_x  & v_y  & v_z \end{matrix} } \right]^T}$ , 
respectively. Its attitude is described by the $Z-Y-X$ Euler angles, which are roll, pitch and yaw angle and denoted by ${\Phi _b} = { \left[ {\begin{matrix} \phi  & \theta  & \psi \end{matrix} } \right]^T}$ . 
So the rotation matrix $^I{R_B}$ is detailed as follow:
\begin{equation}
              {}^I{R_B} = \left[ {\begin{matrix} 
              {c\theta c\psi } & {s\phi s\theta c\psi  - c\phi s\psi } & {c\phi s\theta c\psi  + s\phi s\psi }  \\ 
              {c\theta s\psi } & {s\phi s\theta s\psi  + c\phi c\psi } & {c\phi s\theta s\psi  - s\phi c\psi }  \\ 
              { - s\theta } & {s\phi c\theta } & {c\phi c\theta }  \\  
              \end{matrix} } \right]
          \label{eq18}  
        \end{equation}
where, $c$ and $s$ denote the trigonometric function $cos(.)$ and $sin(.)$, respectively.

As in \cite{jimenez2016modelling}, the coupling effect terms in (\ref{eq12}) and (\ref{eq13}) can be seen as torque and force disturbance of the manipulator to the UAV, so we can separated them from the dynamics. Based on (\ref{eq12}), (\ref{eq13}) and (\ref{eq9}), the dynamics of the hex-rotor manipulator can be expressed as follows:
\begin{equation}
    \begin{cases}
      \dot p_b = {v_b}\\
      \dot v_b=  - \frac{F_t}{m_s}{}^IR{}_B{e_3} + g{e_3} + \frac{F_{dis}}{m_s}\\
      {\dot \Phi }_b=T({\Phi _b}){}^B{\omega _b}\\
      {}^B{{\dot \omega }_b} = I_b^{ - 1}(\tau  - {}^B{\omega _b} \times ({I_b}{}^B{\omega _b}){\text{ + }}{}^B{\tau _{dis}})
    \end{cases} \label{eq19}
\end{equation}
where,  $$T({\Phi _b}){\text{ = }}\left[ {\begin{matrix}
        1 & {\operatorname{s} \phi \tan \theta } & {\operatorname{c} \phi \tan \theta }  \\ 
        0 & {\operatorname{c} \phi } & { - \operatorname{s} \phi }  \\ 
        0 & {\operatorname{s} \phi \sec \theta } & {\operatorname{c} \phi \sec \theta }  \\ 
        \end{matrix} } \right].$$
$I_b$ is the inertia matrix of the hex-rotor referenced to the point $O$ along to body fixed frame axes with respect to $\Sigma _B$. $F_{dis}$ is the force disturbance of manipulator with respect to $\Sigma _I$. $^B \tau _{dis}$ is the torque disturbance of manipulator with respect to $\Sigma _B$. Their detail expansions are as follows:   
\begin{equation}
    \begin{split}
         F_{dis}= & -{m_s}{}^I{R_B}({}^B{\omega _b} \times ({}^B{\omega _b} \times {}^B{r_{oc}}) + {}^B{{\dot \omega }_b} \times {}^B{r_{oc}}  \\ 
         & + 2{}^B{\omega _b} \times {}^B{{\dot r}_{oc}} + {}^B{{\ddot r}_{oc}})
    \end{split} \label{eq20}
\end{equation}
\begin{equation}
    \begin{split}
       & {}^B{\tau _{dis}} = - {}^BI_{man}^o{}^B{{\dot \omega }_b} - {}^B{\omega _b} \times ({}^BI_{man}^o{}^B{\omega _b}) - {}^B\dot I_{man}^o{}^B{\omega _b}  \\ 
       & + {m_s}({}^B{r_{oc}} \times {}^IR_B^{ - 1}(g{e_3} - {{\dot v}_b})) - \frac{m_s^2}{m_{man}}{}^B{r_{oc}} \times {}^B{{\ddot r}_{oc}} \\
       &- \frac{m_s^2}{m_{man}}{}^B{\omega _b} \times ({}^B{r_{oc}} \times {}^B{{\dot r}_{oc}}) 
    \end{split} \label{eq21}
\end{equation}

\textbf{Remark II}: The dynamics of the hex-rotor with a manipulator can be derived into two parts. One part is the dynamics of the hex-rotor itself. The other one is the dynamic coupling part which is denoted as the disturbance of the manipulator, and we call it coupling effect model.

The thrust and torque generated by the rotors of the hex-rotor. They have relationship with the rotational speed of the rotors as the following equations:
\begin{equation}
  \resizebox{0.5\textwidth}{!}{$
    \left[ \begin{matrix}
              {F}_t  \\
              \tau   
            \end{matrix}  \right]
            = \left[ \begin{matrix}
              {{c}_{T}} & {{c}_{T}} & {{c}_{T}} & {{c}_{T}} & {{c}_{T}} & {{c}_{T}}  \\
              -d{{c}_{T}} & d{{c}_{T}} & \frac{1}{2}d{{c}_{T}} & -\frac{1}{2}d{{c}_{T}} & -\frac{1}{2}d{{c}_{T}} & \frac{1}{2}d{{c}_{T}}  \\
              0 & 0 & \frac{\sqrt{3}}{2}d{{c}_{T}} & -\frac{\sqrt{3}}{2}d{{c}_{T}} & \frac{\sqrt{3}}{2}d{{c}_{T}} & -\frac{\sqrt{3}}{2}d{{c}_{T}}  \\
              -{{c}_{\tau }} & {{c}_{\tau }} & -{{c}_{\tau }} & {{c}_{\tau }} & {{c}_{\tau }} & -{{c}_{\tau }}  \\
          \end{matrix} \right]\left[ \begin{matrix}
              {{\omega }_{1}}^{2}  \\
              {{\omega }_{2}}^{2}  \\
              {{\omega }_{3}}^{2}  \\
              {{\omega }_{4}}^{2}  \\
              {{\omega }_{5}}^{2}  \\
              {{\omega }_{6}}^{2}  \\
          \end{matrix} \right] $}
          \label{eq22}
\end{equation}
where, $c_T$ is thrust coefficient. $c_{\tau}$ is torque coefficient. $d$ is the distance from center of rotor to the geometrical center of the hex-rotor and $\omega _i$ ($i=1, 2, \cdots, 6$) is rotational speed of rotor $i$.
\subsection{Control of the Hex-rotor Manipulator}
\begin{figure*}[!t]
    \centering
    \includegraphics[width=0.7\textwidth]{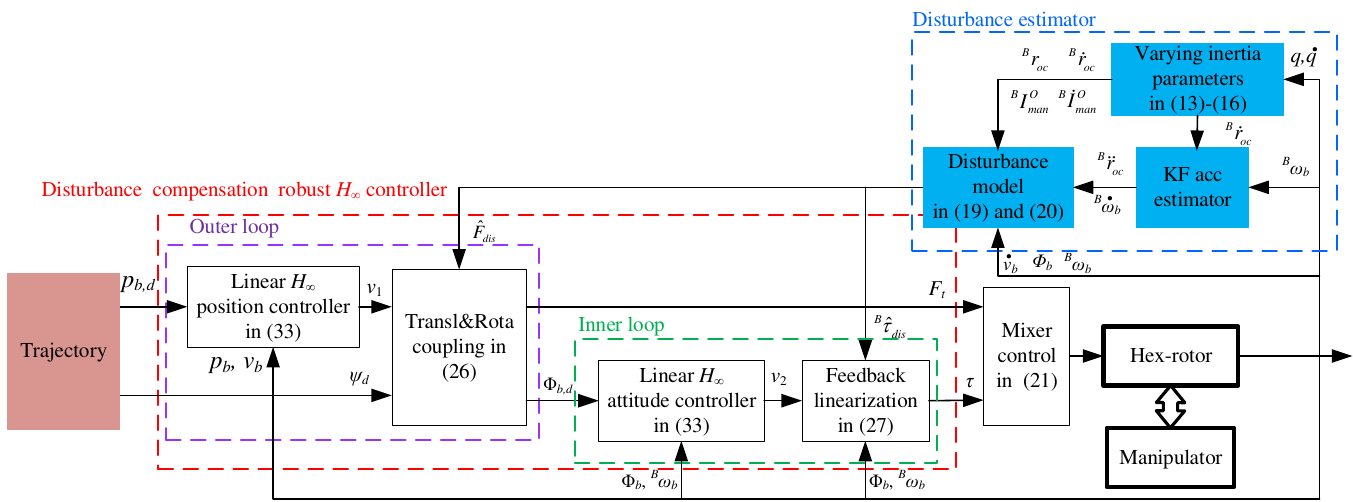}
	\caption{The control structure of the hex-rotor manipulator.}\label{fig_2}
	\vspace{-6pt}
\end{figure*}

The control structure of the hex-rotor in the aerial manipulator system is shown in \mbox{Fig. \ref{fig_2}}. It is composed of a disturbance estimator and a disturbance compensation robust $H_{\infty}$ controller. 

The disturbance estimator is used to estimate the force and torque disturbance based on variable inertia parameters and coupling effect model. The disturbance compensation robust $H_{\infty}$ controller is used to compensate the disturbance of the manipulator and guarantee the stability of the hex-rotor. The robust $H_{\infty}$ controller is based on the hierarchical inner-outer loop control structure. The outer loop is the position control and the inner loop is the attitude control. 
\subsubsection{Disturbance estimator}
From (\ref{eq20}) and (\ref{eq21}), we can know that the force and torque disturbance are the functions of the states UAV and the variable inertia parameters. All these variables are measurable except for ${}^B{\dot \omega _b}$ and ${}^B{\ddot r_{oc}}$. Thus we use a linear estimator (Kalman Filter) to estimate them, as in \mbox{Fig. \ref{fig_2}}. Then, using (\ref{eq20}) and (\ref{eq21}), we can obtain the estimated force disturbance, ${\hat F_{dis}}$ , and the estimated torque disturbance, ${}^B{\hat \tau _{dis}}$ , which are used in the controller to compensate the disturbance of the manipulator.

Considering the estimation error, we can rewrite the system dynamics (\ref{eq19}) with the force and torque estimation as follows: 
\begin{equation}
    \resizebox{0.5\textwidth}{!}{$
    \begin{cases}
      \dot p_b = {v_b}\\
      \dot v_b=  - \frac{F_t}{m_s}{}^IR{}_B{e_3} + g{e_3} +\frac{{\hat F}_{dis}}{m_s}+\frac{\Delta _F}{m_s}\\
      {\dot \Phi }_b=T({\Phi _b}){}^B{\omega _b}\\
      {}^B{{\dot \omega }_b} = I_b^{ - 1}(\tau  - {}^B{\omega _b} \times ({I_b}{}^B{\omega _b})) + I_b^{-1}(\tau  + {}^B{{\hat \tau }_{dis}}) + I_b^{-1}{\Delta _\tau }\\
      \end{cases}
      $} \label{eq23}
\end{equation}
where  ${\Delta _F}={F_{dis}} - {\hat F_{dis}}$ and  ${\Delta _\tau }={}^B{\tau _{dis}} - {}^B{\hat \tau _{dis}}$ are the force and torque disturbance residual error, respectively. 

\textbf{Remark III}: Both ${\Delta _F}$  and  ${\Delta _\tau }$ come from two sources. One is the estimating error, i.e., the error due to the inaccuracy of the estimation algorithm. The other one is the uncertainty. For example, when the aerial manipulator is used in a pick-and-place task, the payload is uncertain. 

\subsubsection{ Translational and rotational dynamics decoupling}
From (\ref{eq23}), we know that the translational and the rotational dynamics are coupled by the rotation matrix $^IR_B$. In order to decouple the two subsystems a virtual control input $\nu _1$ is introduced \cite{kendoul2010guidance}. It is defined as follow:
\begin{equation}
    {\nu _1} =  - \frac{F_t}{m_s}{}^IR{}_{B,d}{e_3} + g{e_3}+\frac{{\hat F}_{dis}}{m_s}
    \label{eq24}
\end{equation}
where, $^IR_{B,d}$ is the rotation matrix determined by the desired attitude angle, denoted by ${\Phi _{b,d}}$ . Substitute (\ref{eq24}) into (\ref{eq23}), the translational dynamics of the hex-rotor can be expressed as follows:
\begin{equation}
    \label{eq25}
    \begin{cases}
      \dot p_b = {v_b}\\
      {\dot v_b} = \nu _1 + \frac{\Delta _F}{m_s} + \delta (e_{\Phi _b})
    \end{cases}
\end{equation}
\begin{equation}
   \label{eq26}
     \delta (e_{\Phi _b}) = \frac{F_t}{m_s}({}^IR{}_{B,d} - {}^IR{}_B){e_3} 
\end{equation}
where, $e_{\Phi _b}={\Phi _b} - {\Phi _{b,d}}$ is the attitude error.  $\delta (e_{\Phi _b})$ is the interconnection term between translational and rotational dynamics after being decoupled by the virtual input $\nu _1$. The virtual input $\nu _1$ is given by the linear $H_{\infty}$ position controller, which will be presented in the next subsection. Then, in order to actuate the outer loop subsystem actually, we need to translate the virtual input $\nu _1$ into thrust, desired roll angle and pitch angle. The vectors of the desired attitude angle and virtual input are defined as:  
$\Phi _{b,d} = {\left[ {\begin{matrix} {\phi _d} & {\theta _d} & {\psi _d} \end{matrix} } \right]^T}$ 
and
${\nu _1}={\left[ {\begin{matrix}   {\nu_{1x}} & {\nu_{1y}} & {\nu_{1z}}\end{matrix} } \right]^T}$ .
Combining (\ref{eq18}) and (\ref{eq24}), the thrust, desired roll angle and desired pitch angle can be got by the following equations:
\begin{equation}
    \label{ee27}
    \begin{cases}
     F_t = \| m_s \nu _1 - m_s g{e_3} - {\hat F}_{dis} \| \\
     \phi _d = \arcsin (\frac{m_s}{F_t}(\nu _{1x} s{\psi _d} - \nu _{1y} c{\psi _d})) \\
     \theta _d = \arctan (\frac{1}{\nu _{1z} - g}(\nu _{1x} c{\psi _d} - \nu _{1y} s{\psi _d}))
    \end{cases}
\end{equation}

Based on the feedback linear theory of the MIMO system , by neglecting the term of $I_b^{-1}{\Delta _\tau }$ , the rotational dynamics in (\ref{eq23}) is feedback linearizable at $\theta  \ne \pi/2$  through the following equation:
\begin{equation}
    \label{eq28}
    \begin{split}
          \tau  = & {I_b}{T^{-1}}({\Phi _b})((-\dot T({\Phi _b}){}^B{\omega _b}  \\ 
         & +T({\Phi _b})I_b^{-1}({}^B{\omega _b} \times ({I_b}{}^B{\omega _b}) + {\nu _2})) - {}^B{{\hat \tau }_{dis}} 
    \end{split}
\end{equation}
where,  $\nu _2$ is the virtual input of the linear system to be used to design controller. 

Through using nominal controller (\ref{eq24}) and (\ref{eq28}), the new system dynamics will be as follows:
\begin{equation}
    \label{eq29}
    \begin{cases}
    \dot v_b = \nu _1 + \frac{\Delta _F}{m_s} + \delta (e_{\Phi _b}) \\
    \ddot \Phi _b= \nu _2
    \end{cases}
\end{equation}
\subsubsection{	$H_{\infty}$ controller design }
With the nominal controller, the new system (\ref{eq29}) become a linear system including interconnection term and disturbances. For the system without disturbances, a linear controller can guarantee the asymptotic stability as in \cite{kendoul2008asymptotic}. In this section, a linear $H_{\infty}$ robust control will be designed to deal with the interconnection term and the disturbance residual error. 

For the hex-rotor, the position and yaw angle are chosen as outputs usually. We use  $p_{b,d}$ to denote the desired position. The system state errors are denoted by $e_{p_b}=p_b-p_{b,d}$, $e_{v_b}=v_b-\dot{p}_{b,d}$, $e_{\Phi_b}=\Phi_b-\Phi_{b,d}$and   $e_{\dot{\Phi}_b}=\dot{\Phi}_b-\dot{\Phi}_{b,d}$. Considering the term of $I_b^{-1}{\Delta _\tau }$  in (\ref{eq23}), the error dynamics of system is as the following equations:
\begin{equation}
\label{eq30} 
          \begin{cases}
            \dot{x}=Ax+Bu+E\delta ({{e}_{\Phi_b}})+D\Delta\\
             y=Cx
          \end{cases}
\end{equation}
where 
        $x=\left[ \begin{matrix}
          e_{p_b}  \\
          e_{p_v}  \\
          e_{\Phi _b}  \\
          e_{\dot{\Phi }_b}  \\
        \end{matrix} \right]$, 
        $y=\left[ \begin{matrix}
          e_x  \\
          e_y  \\
          e_z \\
          e_{\psi }  \\
        \end{matrix} \right]$,
        $u=\left[ \begin{matrix}
          \nu _1-\ddot{p}_{b,d}  \\
          \nu _2-\ddot{\Phi}_{b,d}  \\
        \end{matrix} \right]$,\\
          $A=\left[ \begin{matrix}
            {O_{3\times 3}} & {I_{3\times 3}} & {O_{3\times 3}} & {O_{3\times 3}}  \\
            {O_{3\times 3}} & {O_{3\times 3}} & {O_{3\times 3}} & {O_{3\times 3}}  \\
            {O_{3\times 3}} & {O_{3\times 3}} & {O_{3\times 3}} & {I_{3\times 3}}  \\
            {O_{3\times 3}} & {O_{3\times 3}} & {O_{3\times 3}} & {O_{3\times 3}}  \\
        \end{matrix} \right]$,
        $E=\left[ \begin{matrix}
          {O_{3\times 3}}  \\
          {I_{3\times 3}}  \\
          {O_{3\times 3}}  \\
          {O_{3\times 3}}  \\
        \end{matrix} \right]$,\\
          $B=\left[ \begin{matrix}
            {O_{3\times 3}} & {O_{3\times 3}}  \\
            {I_{3\times 3}} & {O_{3\times 3}}  \\
            {O_{3\times 3}} & {O_{3\times 3}}  \\
            {O_{3\times 3}} & {I_{3\times 3}}  \\
        \end{matrix} \right]$,
        $D=\left[ \begin{matrix}
          {O_{3\times 3}} & {O_{3\times 3}}  \\
          {I_{3\times 3}} & {O_{3\times 3}}  \\
          {O_{3\times 3}} & {O_{3\times 3}}  \\
          {O_{3\times 3}} & {I_{3\times 3}}  \\
        \end{matrix} \right]$,
        \\
        $C=\left[ \begin{matrix}
          {I_{3\times 3}} & {O_{3\times 3}} & {O_{3\times 2}} & {O_{3\times 1}} & {O_{3\times 3}}  \\
          {O_{1\times 3}} & {O_{1\times 3}} & {O_{1\times 2}} & 1 & {O_{1\times 3}}  \\
        \end{matrix} \right]$, 
        \\
        $\Delta =\left[ \begin{matrix}
          \frac{1}{{{m}_{s}}}{{\Delta }_{F}}  \\
          T({{\Phi }_{b}})I_{b}^{-1}{{\Delta }_{\tau }}  \\
        \end{matrix} \right]$.

For the interconnection term $\delta(e_{\Phi_b})$, we have the following Theorem.

\textbf{Theorem I}: 
For the $\delta(e_{\Phi_b})$ in \eqref{eq30}, there exist a constant $\sigma$, satisfy
\begin{equation} \label{deltaLimit}
    \|\delta(e_{\Phi_b})\| \leq \|Fx\|
\end{equation}
where $F = [O_{3\times 3}\quad O_{3\times 3}\quad  \sigma I_{3\times 3}\quad  O_{3\times 3}]$.

The proof of Theorem I is given in the Appendix.

For system (\ref{eq30}), we can design a linear $H_{\infty}$ controller $u=Kx$, and the performance of the linear controller can be
ensured using the following Theorem.

\textbf{Theorem II}: If a linear feedback controller $u=Kx$ and a positive definite symmetric matrix $P$ can be found satisfying the
following inequality\cite{gahinet1994linear}:
\begin{equation}
    \label{eeq31}
    \begin{split}
       & P(A + BK) + {(A + BK)^T}P + \frac{1}{\gamma ^2}PD{D^T}P + \frac{1}{\lambda }PE{E^T}P \\ 
       &  + {C^T}C + \lambda {F^T}F  \leq  0 
    \end{split}
\end{equation}
where, $\lambda$ is a positive constant. Then, system \eqref{eq30} will be finite gain $L_2$-stable from disturbance $\Delta$ to outputs $y$ and the $L_2$ gain is less than or equal to $\gamma$.

\textit{Proof.} Firstly, a Lyapunov function of the closed-loop system is chosen as
\begin{equation*} \label{eq:B_1}
	V(x) = x^TPx
\end{equation*}
Then the differential of $V(x)$ is
\begin{equation*} \label{eq:B_2}
	\begin{aligned}
		\dot{V}(x) & = x^T((A+BK)^T)P + P(A+BK))x + \Delta^TD^TPx                                                                                 \\
		           & + x^TPD\Delta + \delta^T(e_{\Phi_b})E^TPx + x^TPE\delta(e_{\Phi_b})                                                                   \\
		           & = x^T((A+BK)^TP + P(A+BK))x + x^TC^TCx                                                                                       \\
		           & + \frac{1}{\gamma^2}x^TPDD^TPx + \frac{1}{\lambda}x^T PEE^TPx  + \lambda\left\|\delta(e_{\Phi_b})\right\|^2                       \\
		           & - \gamma^2\left\|\Delta - \frac{1}{\gamma^2}D^TPx\right\|^2 - \lambda\left\|\delta(e_{\Phi_b}) - \frac{1}{\lambda}E^TPx\right\|^2 \\
		           & + \gamma^2\left\|\Delta\right\|^2 - \left\|y\right\|^2                                                                       
	\end{aligned}
\end{equation*}
Combining with \eqref{deltaLimit}, we can get
\begin{equation*} \label{eq:B_3}
	\begin{aligned}
		\dot{V}(x) & \leq  x^T((A+BK)^TP + P(A+BK))x + x^TC^TCx                                                             \\
		           & + \frac{1}{\gamma^2}x^TPDD^TPx + \frac{1}{\lambda}x^T PEE^TPx  + \lambda\left\|\delta(e_{\Phi_b})\right\|^2 \\
		           & + \gamma^2\left\|\Delta\right\|^2 - \left\|y\right\|^2                                                 \\
		           & \leq x^T((A+BK)^TP + P(A+BK))x + x^TC^TCx                                                              \\
		           & + \frac{1}{\gamma^2}x^TPDD^TPx + \frac{1}{\lambda}x^T PEE^TPx  + \lambda x^TF^TFx                      \\
		           & + \gamma^2\left\|\Delta\right\|^2 - \left\|y\right\|^2                                                 
	\end{aligned}
\end{equation*}

Thus, if \eqref{eeq31} is satisfied, then
\begin{equation*} \label{eq:B_4}
	\dot{V}(x) \leq \gamma^2\left\|\Delta\right\|^2 - \left\|y\right\|^2
\end{equation*}
which means that, for any $T > 0$, we have
\begin{equation*}
	\begin{aligned}
		  & \int_0^T\left (\left\|y(t)\right\|^2 - \gamma^2\left\|\Delta(t)\right\|^2\right) dt                                        \\
		  & = \int_0^T \left(\left\|y(t)\right\|^2 - \gamma^2\left\|\Delta(t)\right\|^2 + \dot{V}(x(t))\right) dt + V\left(x(T)\right) \\
		  & \leq V\left(x(T)\right)                                                                                                    
	\end{aligned}
\end{equation*}
That is,
\begin{equation*} \label{eq:B_5}
	\int_0^T \left \|y(t)\right\|^2 \leq \int_0^T\gamma^2 \left\|\Delta(t)\right \|^2 dt + V\left(x(T)\right)
\end{equation*}
That means, the system is finite gain $L_2$-stable from disturbance $\Delta$ to outputs $y$ and the $L_2$ gain is less than or equal to $\gamma$.

The gain matrix $K$ can be obtained by solving the following LMI (Linear Matrix Inequality):
\begin{equation}
	\renewcommand\arraystretch{1.5}
	\begin{bmatrix}
		\begin{aligned}
		AX+BW + \\[-2pt]    
		(AX+BW)^T
		\end{aligned}&
		\frac{1}{\gamma}D & \frac{1}{\sqrt{\lambda}}E & (CX)^T & \sqrt{\lambda}(FX)^T \\
		\frac{1}{\gamma}D^T         & -I & O  & O  & O  \\
		\frac{1}{\sqrt{\lambda}}E^T & O  & -I & O  & O  \\
		CX                          & O  & O  & -I & O  \\
		\sqrt{\lambda}FX             & O  & O  & O  & -I 
	\end{bmatrix} \le 0
	\label{eq:LMI}
\end{equation}
where, $X=P^{-1}$. Once $W$ and $X$ obtained, the gain matrix can be got as $K = WX^{-1}$. 
So the virtual input $\nu _1$ and $\nu _2$ is as follows: 
\begin{equation}
    \label{eq32}
    \left[ {\begin{matrix}
          v_1  \\ 
          v_2 \\ 
    \end{matrix} } \right] 
         = Kx + 
    \left[ {\begin{matrix}
         \ddot p_{b,d}  \\ 
         \ddot \Phi _{b,d}  \\ 
     \end{matrix} } \right]
\end{equation}

\section{Experiments}	
In order to validate the proposed coupling effect model and control scheme, two experiments have been conducted and the results will be given and analyzed in this Section. 
\subsection{Experimental Platform}	
\begin{figure}[!t]
    \centering
    \includegraphics[width=7cm]{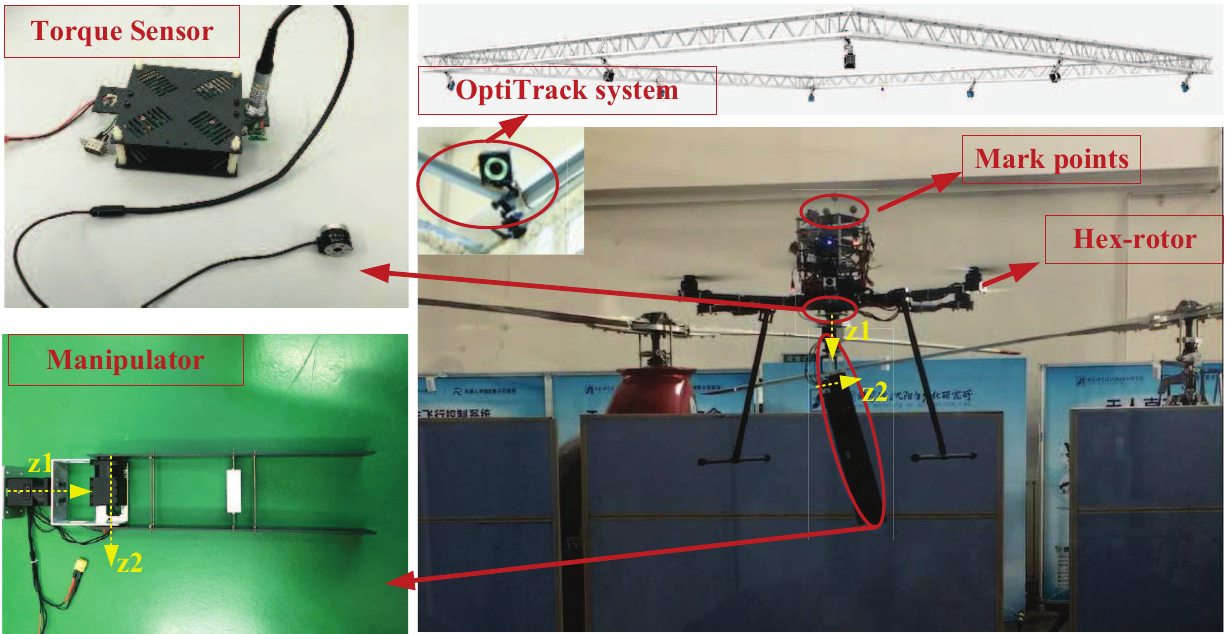}
	\caption{Composition of the hex-rotor manipulator.}\label{fig_3}
	\vspace{-6pt}
\end{figure}

The hex-rotor manipulator we used in the experiments is shown in \mbox{Fig. \ref{fig_3}}. It mainly consists of a hex-rotor UAV, a 2-DoF manipulator and a torque sensor. The torque sensor is installed between the UAV and the manipulator, so that it can directly measure the torque disturbance that the manipulator exerts on the UAV (not used in the controller). The actuators of the manipulator are two Dynamixel Pro M42-10, which can provide high accuracy joint position and joint velocity information. All the experiments are conducted in the OptiTrack system, an indoor motion capture system, which can provide position, velocity and orientation information of the hex-rotor.
\subsection{ Disturbance Measurement Experiment }

In order to validate the coupling effect model proposed in the Section III, we measured the torque disturbance by the torque sensor in the first experiment and compared it with the torque disturbance estimated by the estimator.
 \begin{figure}[!t]
    \centering
    \includegraphics[width=7cm]{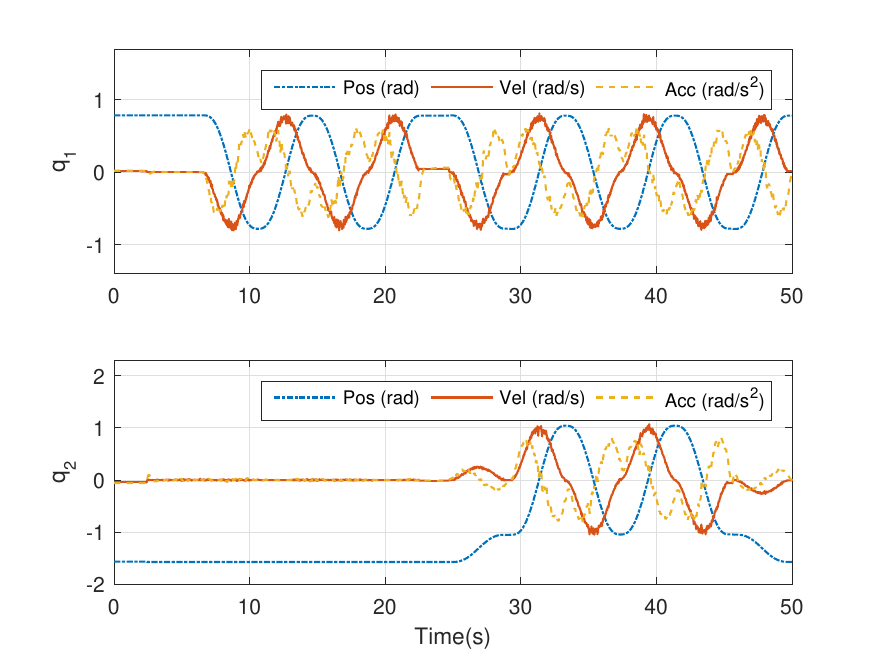}\vspace{-6pt}
	\caption{The joints trajectory of the manipulator in the first experiment.}\label{fig_4}
	\vspace{-6pt}
\end{figure}
At the beginning of the experiment, as shown in \mbox{Fig. \ref{fig_4}}, the first joint of the manipulator swings periodically from $-\pi/4$ to $\pi/4$ while the second joint keeping in $-\pi/2$. Then, at about $t=25s$, the second joint starts to swing periodically for $-\pi/3$ to $\pi/3$. So that the movement of manipulator covers the mainly space of the operational space. In the whole process of the experiment the hex-rotor is flying remotely. 
 \begin{figure}[!t]
    \centering
    \includegraphics[width=7cm]{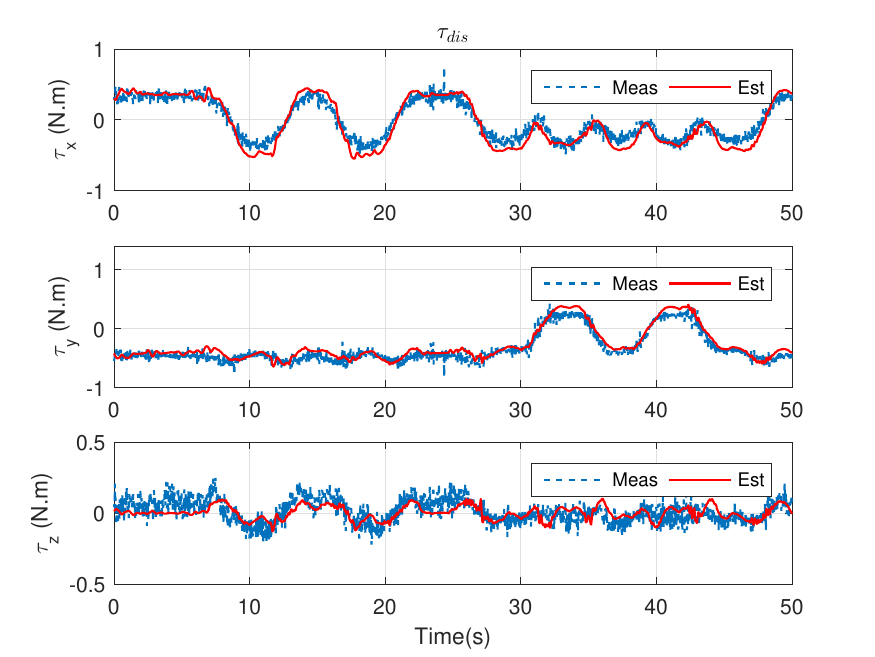}\vspace{-6pt}
	\caption{The measured and estimated torque disturbance.}\label{fig_5}
	\vspace{-6pt}
\end{figure}
The measured and estimated torque disturbance are compared in \mbox{Fig. \ref{fig_5}}. From \mbox{Fig. \ref{fig_5}}, it can be seen that the estimated torque is very close to the measured. To quantitatively show the result, we use the mean absolute percentage deviation (MAPD) to evaluate the errors between the estimated and measured torque disturbance. The index function is defined as follows:
$$J_{index} = \frac{1}{N}\sum\limits_N {\left| \frac{\hat \tau _{dis}(i) - \tau _{dis}(i)}{\tau _{dis}(i)} \right|}  \times 100\% $$	 	
where, $\hat \tau _{dis}(i)$  and  $\tau _{dis}(i)$ are the estimated and measured disturbance at time $i$, respectively. $N$ is the total number of the data.

The results are given in \mbox{Table \ref{table_1}}. The results shown that the estimator is able to steadily output the estimated disturbance torques with a residual errors, about 10\%, of the real disturbances. So coupling effect model can contained the main part of the disturbance, which means that it is feasible to simplify coupling effect model by (\ref{eq9}).
\begin{table}[!t]  \tiny
	\renewcommand{\arraystretch}{1.4}
	\caption{Error mean absolute percentage deviation}
	\centering
	\label{table_1}
	\centering
	\resizebox{\columnwidth}{!}{ 
		\begin{tabular}{l l}
			\hline\hline \\[-3mm] 
			\multicolumn{1}{c}{ Parameter} & \multicolumn{1}{c}{$J_{index}$}\\ \hline \tiny
			{MAPD of $^B \tau _{dis}$ in the $X$ direction \hphantom{1}}  & 9.92 \% \hphantom{1}\\
			{MAPD of $^B \tau _{dis}$ in the $Y$ direction \hphantom{1}}  & 6.59 \% \hphantom{1} \\
			{MAPD of $^B \tau _{dis}$ in the $Z$ direction \hphantom{1}}  & 11.97 \% \hphantom{1}\\
			\hline\hline
		\end{tabular}}
\end{table}
\subsection{ Control Experiment}
In order to validate the proposed control scheme, the aerial manipulator hovering control experiments are conducted. During the experiment, the manipulator swing periodically while the hex-rotor is controlled without and with the disturbance compensation terms, respectively. 

\begin{figure}[!t]
    \centering
    \includegraphics[width=7cm]{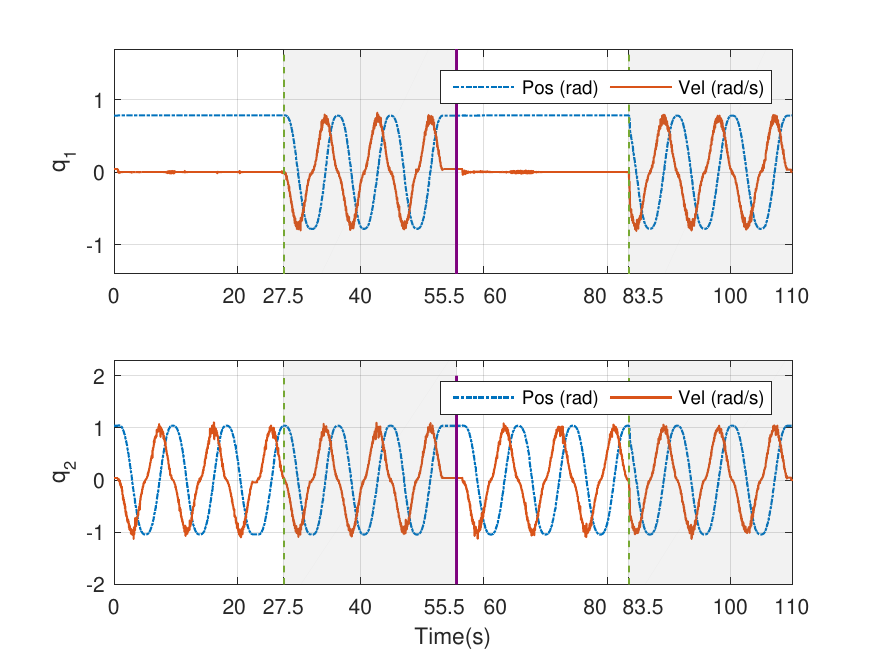}\vspace{-6pt}
	\caption{The joints trajectory of the manipulator in the control experiment (lines in the white and the gray background are the movement of the manipulator in the first and second period, respectively).}\label{fig_6}
	\vspace{-6pt}
\end{figure}

\begin{figure}[!t]
    \centering
    \includegraphics[width=7cm]{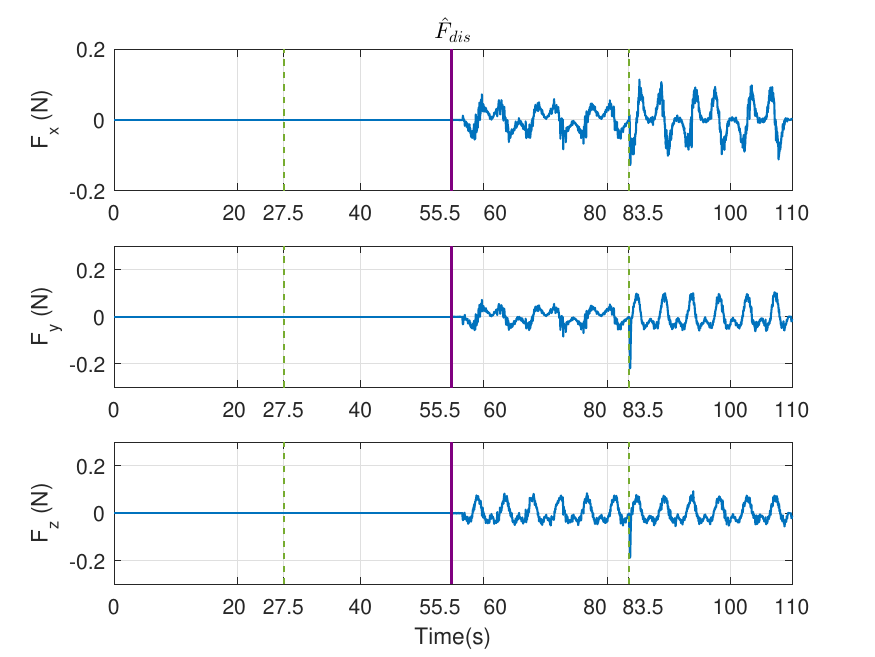}\vspace{-6pt}
	\caption{The estimated force disturbance compensated in the controller.}\label{fig_7}
	\vspace{-6pt}
\end{figure}

\begin{figure}[!t]
    \centering
    \includegraphics[width=7cm]{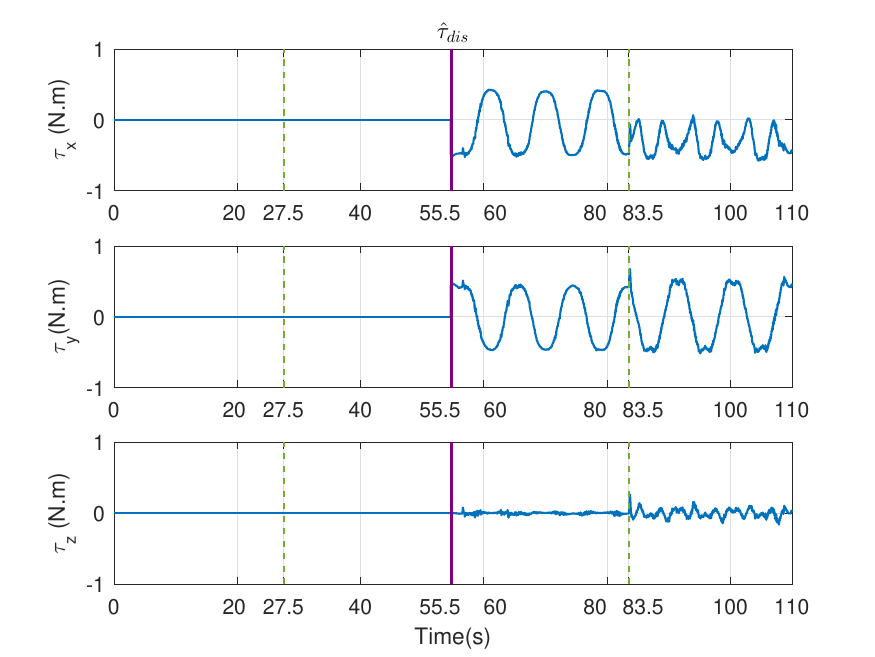}\vspace{-6pt}
	\caption{The estimated torque disturbance compensated in the controller.}\label{fig_8}
	\vspace{-6pt}
\end{figure}

\begin{figure}[!t]
    \centering
    \includegraphics[width=7cm]{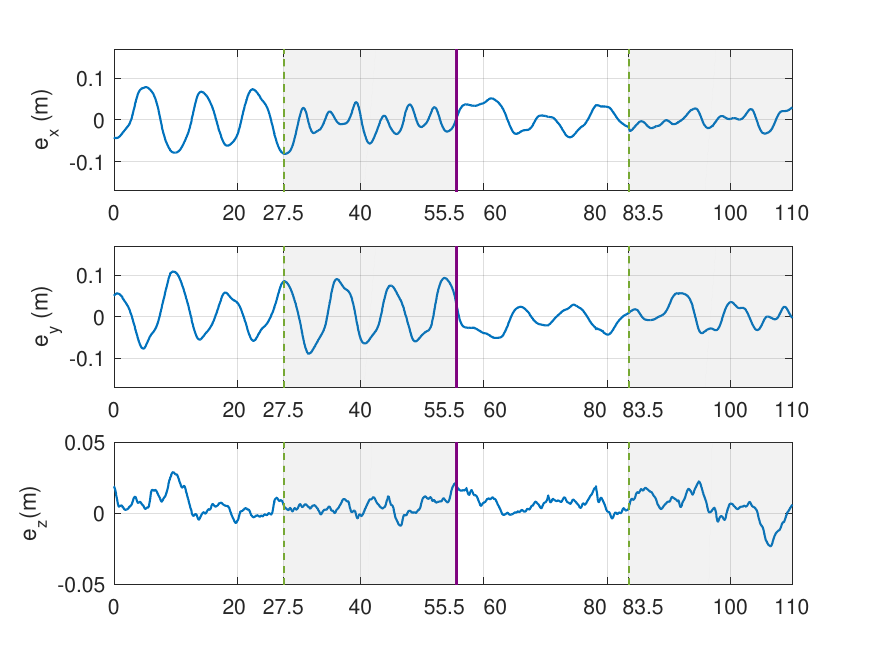}\vspace{-6pt}
	\caption{The position error (at t=0-55.5s and t=55.5-110s hex-rotor is controlled without and with disturbance compensation, respectively).}\label{fig_9}
	\vspace{-6pt}
\end{figure}

\begin{figure}[!t]
    \centering
    \includegraphics[width=7cm]{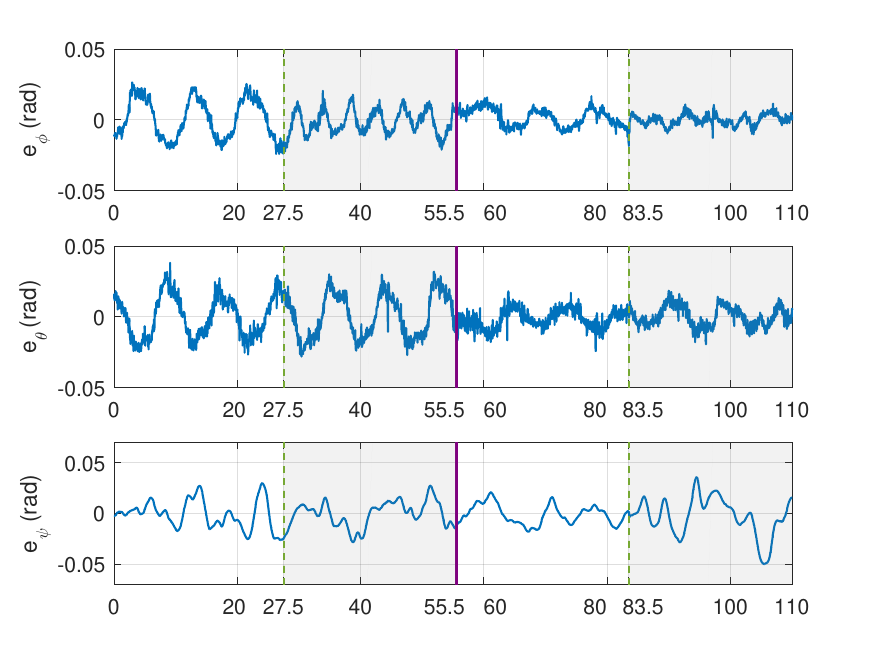}\vspace{-6pt}
	\caption{The attitude error (at t=0-55.5s and t=55.5-110s hex-rotor is controlled without and with disturbance compensation, respectively).}\label{fig_10}
	\vspace{-6pt}
\end{figure}

\begin{figure}[!t]
    \centering
    \includegraphics[width=7cm]{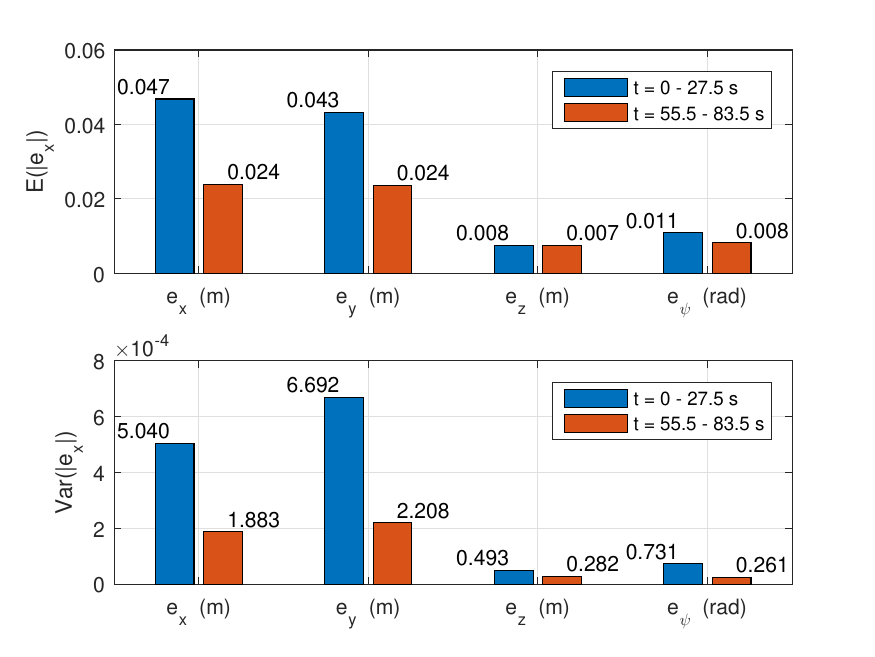}\vspace{-6pt}
	\caption{The mean and variance of absolute error of the system outputs when the manipulator moves in the first period.}\label{fig_11}
	\vspace{-6pt}
\end{figure}

\begin{figure}[!t]
    \centering
    \includegraphics[width=7cm]{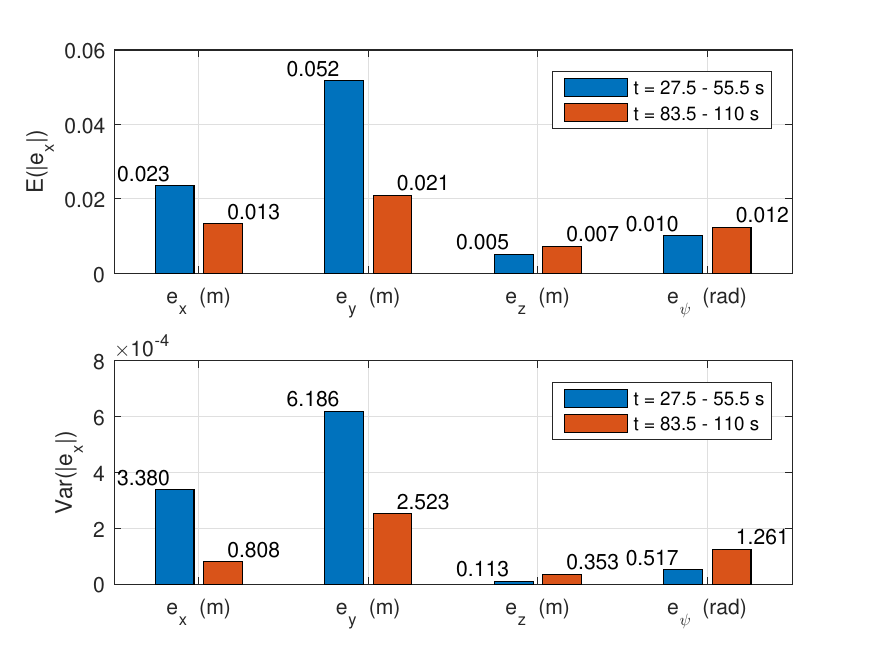}\vspace{-6pt}
	\caption{The mean and variance of absolute error of the system outputs when the manipulator moves in the second period.}\label{fig_12}
	\vspace{-6pt}
\end{figure}

In the experiments, the movement of the manipulator has two periods to generate different disturbances. As shown in \mbox{Fig. \ref{fig_6}}, in the first period, at $t=0-27.5s$ and $t=55.5-83.5s$, the second joint swings periodically from $-\pi/3$ to $\pi/3$ while the first joint fixed in $\pi/3$. In the second period, at $t=27.5-55.5s$ and $t= 83.5-110s$, the first joint swings periodically from $-\pi/4$ to $\pi/4$ while the second joint swings periodically from $-\pi/3$ to $\pi/3$. The estimated force/torque disturbances, compensated in the controller, are shown in \mbox{Fig. \ref{fig_7}} and \mbox{Fig. \ref{fig_8}}. From \mbox{Fig. \ref{fig_7}} and \mbox{Fig. \ref{fig_8}}, we can see that, at $t=0-55.5s$, the hex-rotor is controlled without disturbance compensation and then, at $t=55.5-110s$, the disturbance compensation terms are added in the controller. The results of the position and attitude errors are shown in \mbox{Fig. \ref{fig_9}} and \mbox{Fig. \ref{fig_10}}. The mean and variance of absolute error of the system outputs are shown in the \mbox{Fig. \ref{fig_11}} and \mbox{Fig. \ref{fig_12}}. (The experimental video is available at the following website: {\color{blue}\underline{\url{https://v.youku.com/v_show/id_XMzc5MjY4OTg2OA==.html}}}).

As shown in \mbox{Fig. \ref{fig_9}}, \mbox{Fig. \ref{fig_10}} 
and \mbox{Fig. \ref{fig_11}}, when the manipulator moves in the first 
period (lines in the white background in \mbox{Fig. \ref{fig_9}} and \mbox{Fig. \ref{fig_10}}), the position and attitude error of the hex-rotor controlled 
with disturbance compensation (at $t=55.5-83.5s$) are  reduced 
obviously contrasting to those of the hex-rotor 
controlled without disturbance compensation (at $t=0-27.5s$). 
When the manipulator moves in the second period 
(lines in the gray background in \mbox{Fig. \ref{fig_9}} and \mbox{Fig. \ref{fig_10}}), 
the results are similar. The experimental results can be summarized 
as follows:
\begin{enumerate}[1)]
\item The free moving manipulator causes degradation of the control performance mainly in $X$ and $Y$ directions of the system outputs. In the whole experiment, the mean and variance of absolute error in $Z$ direction are so small that they are hardly to be decreased by the compensation terms.
\item	The disturbance compensation terms can compensate the disturbance of the manipulator and improve the control accuracy of the hex-rotor outputs mainly in $X$ and $Y$ directions. Specifically, the mean and variance of absolute error of the system outputs are decreased by half approximately.
\end{enumerate}

\section{Conclusion}
In this paper, we proposed an UAV dynamic model which include the disturbance that the manipulator effect on the UAV. The disturbance is described by the variable inertia parameters of the aerial manipulator system. Then, based on the dynamic model, a disturbance estimator and a disturbance compensation robust $H_{\infty}$ controller are designed to make the UAV to fly steady when the manipulator is manipulating. Experiments results show that the proposed control scheme can make the UAV hovering more steady than the one without the disturbance compensation terms. 

\section*{Appendix}

\textit{Proof of Therom I}

Similar with the proof in \cite{kendoul2008asymptotic}, define $h(e_{\Phi_b}) = [h_x,h_y,h_z]^T = ( {}^IR_{B,d} -{}^I R_B)e_3$, and combine with \eqref{eq18}, we can obtain
\begin{equation} \label{eq:A_1}
	\left\{
	\begin{aligned}
		h_x & = c\phi s\theta c\psi + s\phi s\psi - (c\phi_{d}s\theta_{d}c\phi_{d} + s\phi_d s\psi_d) \\
		h_y & = c\phi s\theta s\psi - s\phi c\psi - (c\phi_{d}s\theta_{d}s\phi_{d} + s\phi_d c\psi_d) \\
		h_z & = c\phi c\theta - c\phi_d c\theta_d                                                     
	\end{aligned}
	\right.
\end{equation}
For $h_x$ in \eqref{eq:A_1}, replacing $[\phi, \theta, \psi] ^T$ by $[\phi_d + e_\phi, \theta_d + e_\theta, \phi_d + e_\psi]^T$, 
and combining with the following trigonometric inequalities: 
\begin{equation} 
	\left\{
	\begin{aligned}
		\sin(a+b) & = \sin(a)+\sin(\frac{b}{2})\cos(a+\frac{b}{2}) \\
		\cos(a+b) & = \cos(a)-\sin(\frac{b}{2})\sin(a+\frac{b}{2}) \\
    |\sin(a)| & \leq 1\\
    |\cos(a)| & \leq 1                                                     
	\end{aligned}
	\right.
\end{equation}
We can obtain
\begin{equation} \label{eq:A_3}
	\begin{aligned}
		|h_x| & \leq 2\left|s\frac{e_\phi}{2}\right| + \left|s\frac{e_\theta}{2}\right| + 2\left|s\frac{e_\psi}{2}\right| + \left|s\frac{e_\phi}{2}\right|\cdot\left|s\frac{e_\theta}{2}\right| \\
		      & + \left|s\frac{e_\theta}{2}\right|\cdot\left|s\frac{e_\psi}{2}\right| + 2\left|s\frac{e_\phi}{2}\right|\cdot\left|s\frac{e_\psi}{2}\right|                                      \\
		      & + \left|s\frac{e_\phi}{2}\right|\cdot\left|s\frac{e_\theta}{2}\right|\cdot\left|s\frac{e_\psi}{2}\right|                                                                        
	\end{aligned}
\end{equation}
Combining with the following inequalities into \eqref{eq:A_3}
\begin{equation} \label{eq:A_4}
	\left\{
	\begin{aligned}
		  & |a|\cdot|b| \leq \frac{1}{2}(|a| + |b|),~for~|a|,~|b| \leq 1                    \\
		  & |a|\cdot|b|\cdot|c| \leq \frac{1}{3}(|a| + |b| + |c|),~for~|a|,~|b|,~|c| \leq 1 \\
		  & |\sin(a)| \leq |a|                                                              
	\end{aligned}
	\right.
\end{equation}
Thus, we can obtain
\begin{equation} \label{eq:A_5}
	|h_x| \leq \frac{10}{3}\left|s\frac{e_\phi}{2}\right| + \frac{7}{3}\left|s\frac{e_\theta}{2}\right| + \frac{10}{3}\left|s\frac{e_\psi}{2}\right| \leq \frac{5}{3}(|e_\phi| + |e_\theta| + |e_\psi|)
\end{equation}
Similarly, the property of $h_y$ and $h_z$ is
\begin{equation} \label{eq:A_6}
	\left\{
	\begin{aligned}
		|h_y| & \leq \frac{5}{3}(|e_\phi| + |e_\theta| + |e_\psi|) \\
		|h_z| & \leq \frac{3}{4}(|e_\phi| + |e_\theta|)            
	\end{aligned}
	\right.
\end{equation}
With \eqref{eq:A_5} and \eqref{eq:A_6}, the norm of $h(e_{\Phi_b})$ satisfies
\begin{equation} \label{eq:A_7}
	\|h(e_{\Phi_b})\| = \sqrt{h_x^2 + h_y^2 + h_z^2} \leq k_1\|e_{\Phi_b}\|
\end{equation}
where $k_1 \leq \sqrt{13}$. For hex-rotor, we can assume the maximum thrust is $k_2$, thus
\begin{equation} \label{eq:A_8}
	\begin{aligned}
		\left\|\delta(e_{\Phi_b}) \right\| &\leq \left\|\frac{F_t}{m_s}\right\| \cdot \|h(e_{\Phi_b})\| \leq \frac{k_1 k_2}{_s} \|e_{\Phi_b}\| \\
		& = \left\|
		\begin{bmatrix}
		O_{3\times 3} & O_{3\times 3} & \sigma I_{3\times 3} & O_{3\times 3} 
		\end{bmatrix} x \right\|
	\end{aligned}
\end{equation}
where $\sigma = \frac{k_1 k_2}{m_s}$. 



\bibliographystyle{IEEEtranTIE}
\bibliography{BIB_19-TIE-1923}\ 

\begin{thebibliography}{10}
\providecommand{\url}[1]{#1}
\csname url@samestyle\endcsname
\providecommand{\newblock}{\relax}
\providecommand{\bibinfo}[2]{#2}
\providecommand{\BIBentrySTDinterwordspacing}{\spaceskip=0pt\relax}
\providecommand{\BIBentryALTinterwordstretchfactor}{4}
\providecommand{\BIBentryALTinterwordspacing}{\spaceskip=\fontdimen2\font plus
\BIBentryALTinterwordstretchfactor\fontdimen3\font minus \fontdimen4\font\relax}
\providecommand{\BIBforeignlanguage}[2]{{%
\expandafter\ifx\csname l@#1\endcsname\relax
\typeout{** WARNING: IEEEtran.bst: No hyphenation pattern has been}%
\typeout{** loaded for the language `#1'. Using the pattern for}%
\typeout{** the default language instead.}%
\else
\language=\csname l@#1\endcsname
\fi
#2}}
\providecommand{\BIBdecl}{\relax}
\BIBdecl

\bibitem{khamseh2018aerial}
H.~B. Khamseh, F.~Janabi-Sharifi, and A.~Abdessameud, ``Aerial manipulation-a literature survey,'' \emph{Robot. Auton. Syst.}, vol. 107, pp. 221--235, Sep. 2018.

\bibitem{lee2017estimation}
H.~Lee and H.~J. Kim, ``Estimation, control, and planning for autonomous aerial transportation,'' \emph{{IEEE} Trans. Ind. Electron}, vol.~64, no.~4, pp. 3369--3379, Apr. 2017.

\bibitem{thomas2014toward}
J.~Thomas, G.~Loianno, J.~Polin, K.~Sreenath, and V.~Kumar, ``Toward autonomous avian-inspired grasping for micro aerial vehicles,'' \emph{Bioinspir. Biomim.}, vol.~9, no.~2, p. 025010, 2014.

\bibitem{lippiello2012cartesian}
V.~Lippiello and F.~Ruggiero, ``Cartesian impedance control of a uav with a robotic arm,'' \emph{IFAC Proc.}, vol.~45, no.~22, pp. 704--709, Jan. 2012.

\bibitem{lippiello2012exploiting}
V.~Lippiello and F.~Ruggiero, ``Exploiting redundancy in cartesian impedance control of uavs equipped with a robotic arm,'' in \emph{Proc. IEEE/RSJ Int. Conf. Intell. Robots Syst.}, pp. 3768--3773.\hskip 1em plus 0.5em minus 0.4em\relax IEEE, 2012.

\bibitem{kim2016vision}
S.~Kim, H.~Seo, S.~Choi, and H.~J. Kim, ``Vision-guided aerial manipulation using a multirotor with a robotic arm,'' \emph{IEEE/ASME Trans. Mechatron.}, vol.~21, no.~4, pp. 1912--1923, Aug. 2016.

\bibitem{yang2014dynamics}
H.~Yang and D.~Lee, ``Dynamics and control of quadrotor with robotic manipulator,'' in \emph{Proc. IEEE Int. Conf. Robot. Autom.}, pp. 5544--5549.\hskip 1em plus 0.5em minus 0.4em\relax IEEE, 2014.

\bibitem{kobilarov2014nonlinear}
M.~Kobilarov, ``Nonlinear trajectory control of multi-body aerial manipulators,'' \emph{J. Intell. Robot. Syst.}, vol.~73, no. 1-4, pp. 679--692, Jan. 2014.

\bibitem{caccavale_adaptive_2014}
F.~Caccavale, G.~Giglio, G.~Muscio, and F.~Pierri, ``Adaptive control for {UAVs} equipped with a robotic arm,'' \emph{IFAC Proc.}, vol.~47, no.~3, pp. 11\,049--11\,054, Jan. 2014.

\bibitem{arleo_control_2013}
G.~Arleo, F.~Caccavale, G.~Muscio, and F.~Pierri, ``Control of quadrotor aerial vehicles equipped with a robotic arm,'' in \emph{Proc. 21st Medit. Conf. Control Autom.}, pp. 1174--1180.\hskip 1em plus 0.5em minus 0.4em\relax Chania: IEEE, Jun. 2013.

\bibitem{kendoul2008asymptotic}
F.~Kendoul, I.~Fantoni, R.~Lozano \emph{et~al.}, ``Asymptotic stability of hierarchical inner-outer loop-based flight controllers,'' \emph{IFAC Proc.}, vol.~41, no.~2, pp. 1741--1746, 2008.

\bibitem{kondak2013closed}
K.~Kondak, K.~Krieger, A.~Albu-Schaeffer, M.~Schwarzbach, M.~Laiacker, I.~Maza, A.~Rodriguez-Castano, and A.~Ollero, ``Closed-loop behavior of an autonomous helicopter equipped with a robotic arm for aerial manipulation tasks,'' \emph{Int. J. Adv. Robot. Syst.}, vol.~10, no.~2, p. 145, Feb. 2013.

\bibitem{kim2018robust}
S.~Kim, S.~Choi, H.~Kim, J.~Shin, H.~Shim, and H.~J. Kim, ``Robust control of an equipment-added multirotor using disturbance observer,'' \emph{IEEE Trans. Control Syst. Technol.}, vol.~26, no.~4, pp. 1524--1531, 2018.

\bibitem{khalifa2017new}
A.~Khalifa and M.~Fanni, ``A new quadrotor manipulation system: Modeling and point-to-point task space control,'' \emph{Int. J. Control Autom. Syst.}, vol.~15, no.~3, pp. 1434--1446, 2017.

\bibitem{fanni2017new}
M.~Fanni and A.~Khalifa, ``A new 6-dof quadrotor manipulation system: design, kinematics, dynamics, and control,'' \emph{IEEE/ASME Trans. Mechatron.}, vol.~22, no.~3, pp. 1315--1326, 2017.

\bibitem{jimenez2016modelling}
A.~Jimenez-Cano, G.~Heredia, M.~Bejar, K.~Kondak, and A.~Ollero, ``Modelling and control of an aerial manipulator consisting of an autonomous helicopter equipped with a multi-link robotic arm,'' \emph{Proc. Inst. Mech. Eng. G}, vol. 230, no.~10, pp. 1860--1870, 2016.

\bibitem{orsag2017dexterous}
M.~Orsag, C.~Korpela, S.~Bogdan, and P.~Oh, ``Dexterous aerial robots-mobile manipulation using unmanned aerial systems,'' \emph{IEEE Trans. Robot.}, vol.~33, no.~6, pp. 1453--1466, Dec. 2017.

\bibitem{zhang2018grasp}
G.~Zhang, Y.~He, B.~Dai, F.~Gu, L.~Yang, J.~Han, G.~Liu, and J.~Qi, ``Grasp a moving target from the air: System \& control of an aerial manipulator,'' in \emph{Proc. IEEE Int. Conf. Robot. Autom.}, pp. 1681--1687.\hskip 1em plus 0.5em minus 0.4em\relax IEEE, 2018.

\bibitem{wittenburg2007dynamics}
J.~Wittenburg, \emph{Dynamics of multibody systems}.\hskip 1em plus 0.5em minus 0.4em\relax Springer Science \& Business Media, 2007.

\bibitem{kendoul2010guidance}
F.~Kendoul, Z.~Yu, and K.~Nonami, ``Guidance and nonlinear control system for autonomous flight of minirotorcraft unmanned aerial vehicles,'' \emph{J. Field Robot.}, vol.~27, no.~3, pp. 311--334, May. 2010.

\bibitem{gahinet1994linear}
P.~Gahinet and P.~Apkarian, ``A linear matrix inequality approach to $h_{\infty}$ control,'' \emph{Int. J. Robust Nonlinear Control}, vol.~4, no.~4, pp. 421--448, 1994.

\end{thebibliography}

{
\vspace{-1cm}
\begin{IEEEbiography}[{\includegraphics[width=1in,height=1.25in,clip,keepaspectratio]{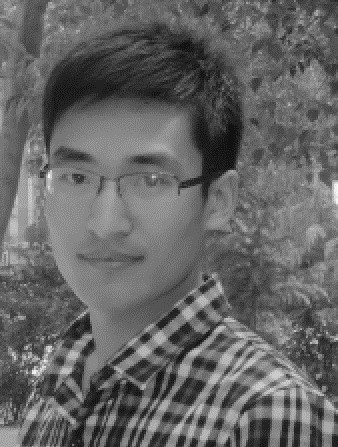}}]
{Guangyu Zhang} received the B.E. degree in vehicle engineering from Lanzhou Jiaotong University, Lanzhou, China, in 2013. He is currently working toward the Ph.D. degree in pattern recognition and intelligent system at the Shenyang Institute of Automation, Chinese Academy of Sciences, Shenyang, China. His main research interests include dynamic modeling, control and motion planning of aerial manipulator.
\end{IEEEbiography}

\vspace{-1cm}
\begin{IEEEbiography}[{\includegraphics[width=1in,height=1.25in,clip,keepaspectratio]{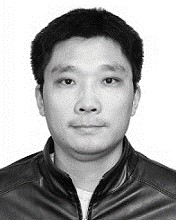}}]
{Yuqing He} (M'12) was born in Weihui, China, in 1980. He received the B.E. degree in engineering and automation from Northeastern University, Qinhuangdao, China, in 2002, and the Ph.D. degree in pattern recognition and intelligent system from the Shenyang Institute of Automation, Chinese Academy of Sciences, Shenyang, China, in 2008. 

He is currently a Professor at the State Key Laboratory of Robotics, Shenyang Institute of Automation, Chinese Academy of Sciences, Shenyang. In 2012, he was a Visiting Researcher at the Institute for Automatic Control Theory, Technical University of Dresden, Germany. His main research interests include nonlinear estimation, control, and cooperation of multiple robots.
\end{IEEEbiography}

\vspace{-1cm}
\begin{IEEEbiography}[{\includegraphics[width=1in,height=1.25in,clip,keepaspectratio]{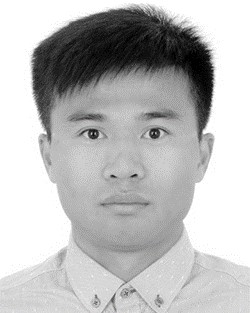}}]
{Bodai} was born in Huainan, China, in 1993. He received the B.E. degree in engineering and automation from University of Science and Technology Beijing, Beijing, China, in 2014. He is currently working toward the Ph.D. degree in pattern recognition and intelligent system at the Shenyang Institute of Automation, Chinese Academy of Sciences, Shenyang, China. His main research interests include control and motion planning of aerial vehicle.
\end{IEEEbiography}

\vspace{-1cm}
\begin{IEEEbiography}[{\includegraphics[width=1in,height=1.25in,clip,keepaspectratio]{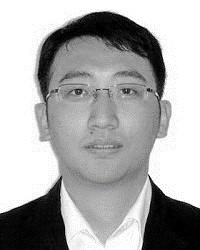}}]
{Feng Gu} was born in Shenyang, China, in 1982. He received the B.E. and M.E. degrees in system engineering from Nanjing University of Science and Technology, Nanjing, China, in 2005 and 2007, respectively, and the Ph.D. degree in pattern recognition and intelligent systems from the Shenyang Institute of Automation, Chinese Academy of Sciences, Shenyang, China, in 2011.

He is currently an Associate Professor at the State Key Laboratory of Robotics, Shenyang Institute of Automation, Chinese Academy of Sciences, Shenyang. His current research interests include set-theory-based estimation and control, cooperation of multiple robots, and research work on mobile robots. 
\end{IEEEbiography}

\vspace{-1cm}
\begin{IEEEbiography}[{\includegraphics[width=1in,height=1.25in,clip,keepaspectratio]{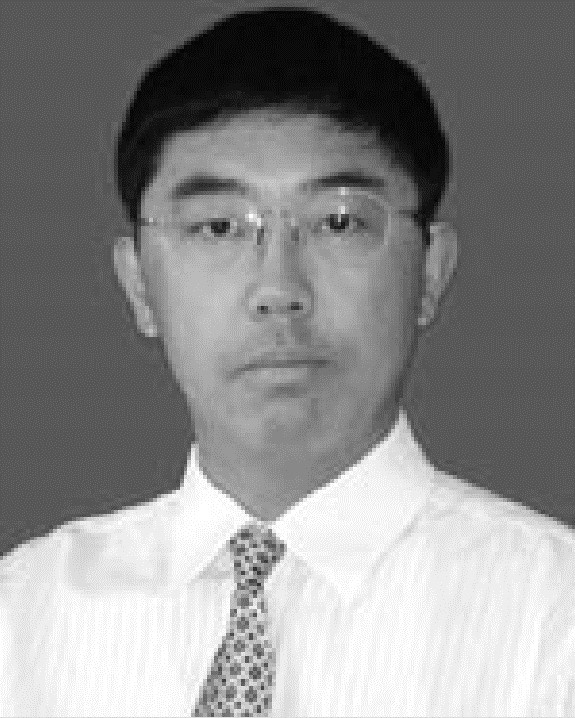}}]
{Jianda Han} (M'05) was born in Liaoning, China, in 1968. He received the Ph.D. degree from Harbin Institute of Technology, Harbin, China, in 1998. 

He is currently a Professor and the ViceDirector of the State Key Laboratory of Robotics, Shenyang Institute of Automation, Chinese Academy of Sciences, Shenyang, China. His research interests include nonlinear estimation and control, robotics, and mechatronics systems.
\end{IEEEbiography}

\vspace{-1cm}
\begin{IEEEbiography}[{\includegraphics[width=1in,height=1.25in,clip,keepaspectratio]{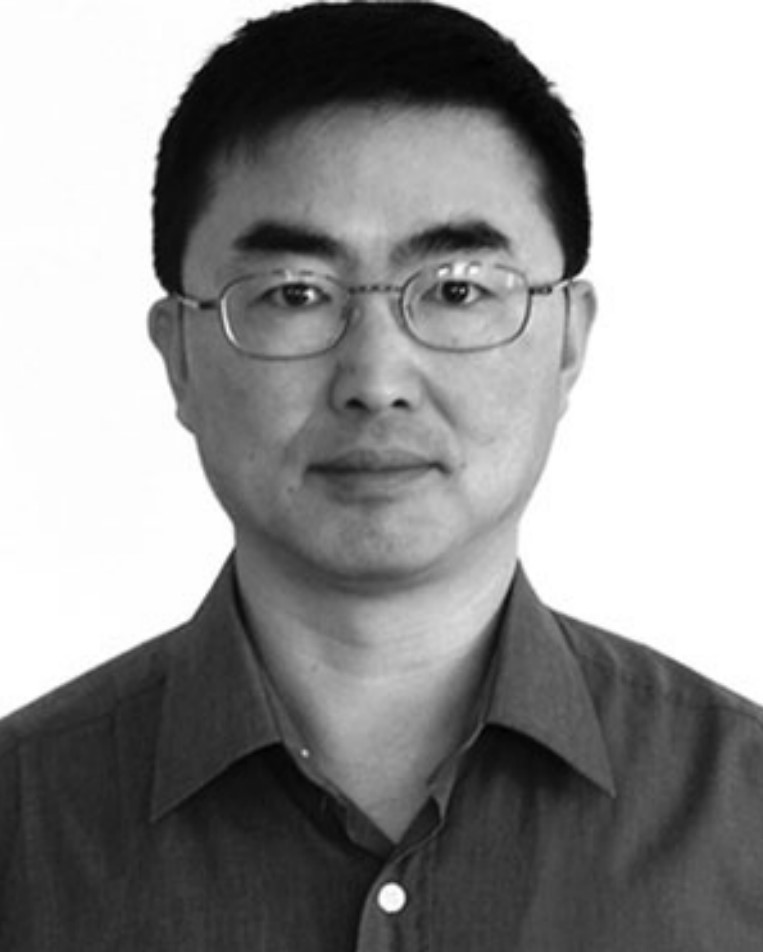}}]
{Guangjun Liu} (M'99-SM'08) received the B.E. degree from University of Science and Technology of China, Hefei, China, in 1984; the M.E. degree from Chinese Academy of Sciences, Shenyang Institute of Automation, Shenyang, China, in 1987; and the Ph.D. degree from University of Toronto, Toronto, ON, Canada, in 1996.

From 1997 to 1999, he was a Systems Engineer and a Design Lead for Honeywell Aerospace Canada, where he was involved in the Boeing X-32 program. In 1996, he was a Postdoctoral Fellow with Massachusetts Institute of Technology, Cambridge, MA, USA. He is currently a Professor and the Canada Research Chair in Control Systems and Robotics with the Department of Aerospace Engineering, Ryerson University, Toronto, ON, Canada. His research interests include control systems and robotics, particularly in modular and reconfigurable robots, mobile manipulators, and aircraft systems.

Dr. Liu is a former Technical Editor of IEEE/ASME TRANSACTIONS ON MECHATRONICS and a licensed Member of the Professional Engineers of Ontario, Canada.
\end{IEEEbiography}
}

\end{document}